\documentclass{article}
\usepackage{subcaption}
\usepackage{spconf,amsmath,graphicx}
\usepackage{tikz}
\usepackage[hyphens]{url}

\usepackage{booktabs}

\title{Residual Attention based Network for Hand Bone Age Assessment}
\name{Author(s) Name(s)}
\address{Author Affiliation(s)}
\name{E. Wu$^{1,\star}$, B. Kong$^{3,\star}$, X. Wang$^{2,\dagger}$, J. Bai$^{2}$, Y. Lu$^{2}$, F. Gao$^{2}$, S. Zhang$^{3}$, K. Cao$^{2}$, Q. Song$^{2}$, S. Lyu$^{4}$, Y. Yin$^{2,\dagger}$}
\address{
$^{1}$Cornell University, Ithaca, NY, USA\\
$^{2}$CuraCloud Corporation, Seattle, WA, USA\\
$^{3}$Department of Computer Science, UNC Charlotte, Charlotte, NC, USA\\
$^{4}$Department of Computer Science, University at Albany, State University at New York, NY, USA}
\begin{document}
%
\maketitle
\begin{abstract}
Computerized automatic methods have been employed to boost the productivity as well as objectiveness of hand bone age assessment. These approaches make predictions according to the whole X-ray images, which include other objects that may introduce distractions. Instead, our framework is inspired by the clinical workflow (Tanner-Whitehouse) of hand bone age assessment, which focuses on the key components of the hand. The proposed framework is composed of two components: a Mask R-CNN subnet of pixelwise hand segmentation and a residual attention network for hand bone age assessment. The Mask R-CNN subnet segments the hands from X-ray images to avoid the distractions of other objects (e.g., X-ray tags). The hierarchical attention components of the residual attention subnet force our network to focus on the key components of the X-ray images and generate the final predictions as well as the associated visual supports, which is similar to the assessment procedure of clinicians. We evaluate the performance of the proposed pipeline  on the RSNA pediatric bone age dataset\footnote{http://rsnachallenges.cloudapp.net/competitions/4} and the results demonstrate its superiority over the previous methods.

\renewcommand{\thefootnote}{\fnsymbol{footnote}}
\footnotetext[1]{Indicates equal contribution}
\footnotetext[2]{Corresponding authors}

\end{abstract}
\section{Introduction}
\label{sec:intro}
Although bone age assessment is crucial for the evaluation of the status of many diseases, the clinical procedure remains almost the same since the seminal work of Greulich and Pyle (G\&P)~\cite{greulich1959radiographic}, in which doctors compare an X-ray of the hand that includes the fingers and wrist with an atlas of X-rays to determine the bone age. Since this process requires a significant amount of time and expertise to scrutinize X-rays, it is extremely tedious and error-prone. Additionally, it introduces the intra- and inter-observer variabilities.

Computer-aided diagnosis (CAD) bone age assessment approaches have been developed to address the above issues. For instance, BoneXpert~\cite{thodberg2009bonexpert} with hand-engineered image processing approaches has been developed and approved for use in various countries. However, significant variations of size, shape, and mineralization exist in X-ray images. These often lead to the inaccurate prediction of age. 

\begin{figure}[h]
\begin{center}
\includegraphics[width=0.95\linewidth]{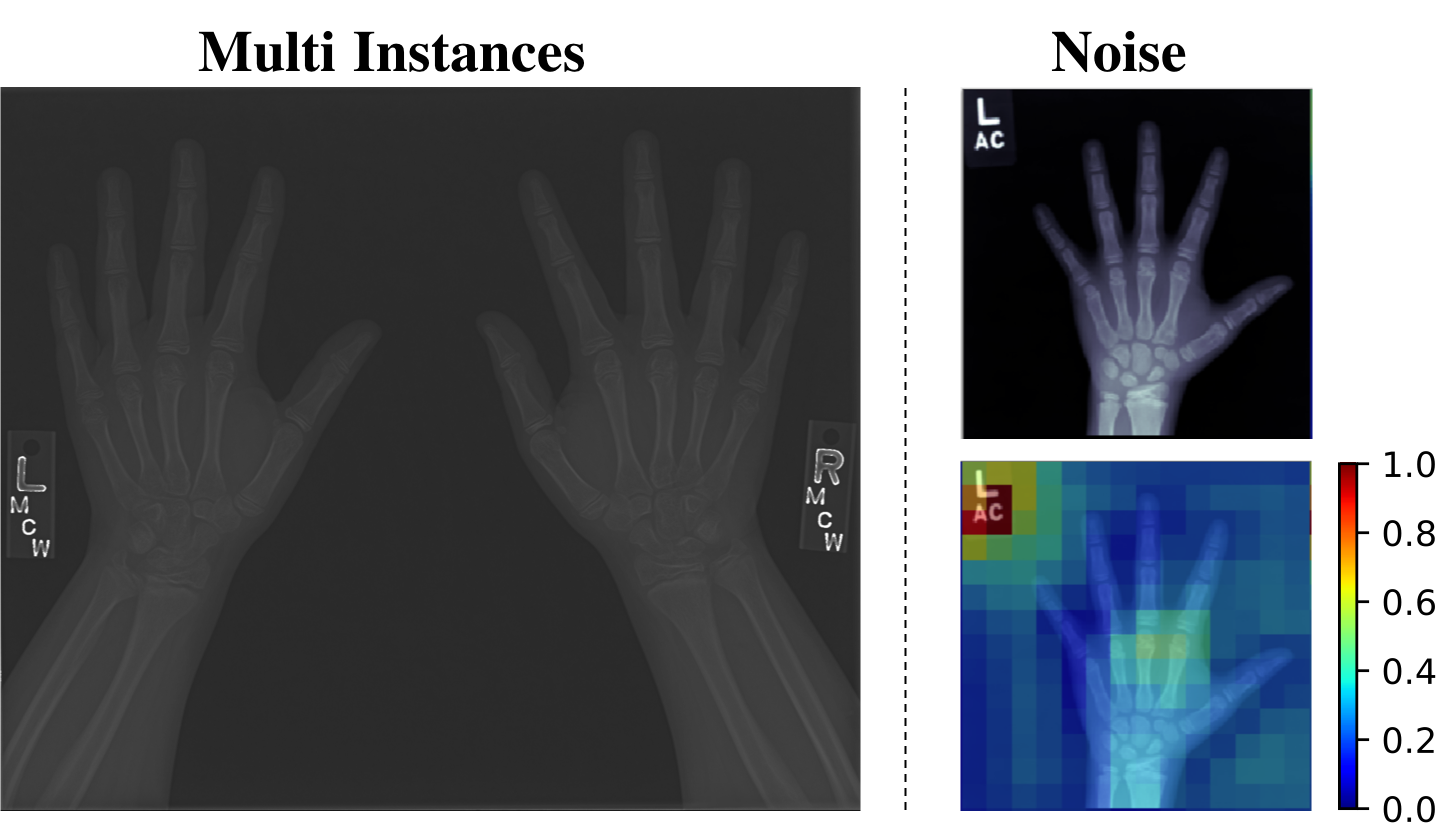} 
\end{center}
~\vspace{-3.0em}
\caption{Examples of hand X-ray images with multi-instances (left), and incorrect attention on the noise background (right).}
\label{fig:hardexample} 
~\vspace{-1.5em}
\end{figure}
Recently, deep learning with hierarchical structures has been adopted as a methodology of choice for many medical image analysis problems~\cite{kong2017cancer,kong2018invasive}. It has demonstrated its superior performance over other machine learning techniques. Several works have adopted deep learning to determine bone age according to the whole X-ray image. For instance, Spampinato et al.~\cite{spampinato2017deep} employed BoNet with multiple convolution, deformation, and fully connected layers to automatically learn to predict bone age. Nevertheless, other objects (e.g., X-ray tags and annotation markers) besides hands also exist in X-ray images. For instance, the right of Fig.~\ref{fig:hardexample} shows a hand X-ray image and its corresponding attention map (large value indicates more importance to the final prediction) generated from the deep learning model trained the same way as in~\cite{spampinato2017deep}. Obviously, these objects act as noises that distract the network to other unimportant regions of the image, thereby yielding a suboptimal result. Additionally, some images have more than one instance of hand (left of Fig.~\ref{fig:hardexample}), which makes the prediction more challenging.

\begin{figure*}[htb]
\begin{center}
\includegraphics[width=0.95\textwidth]{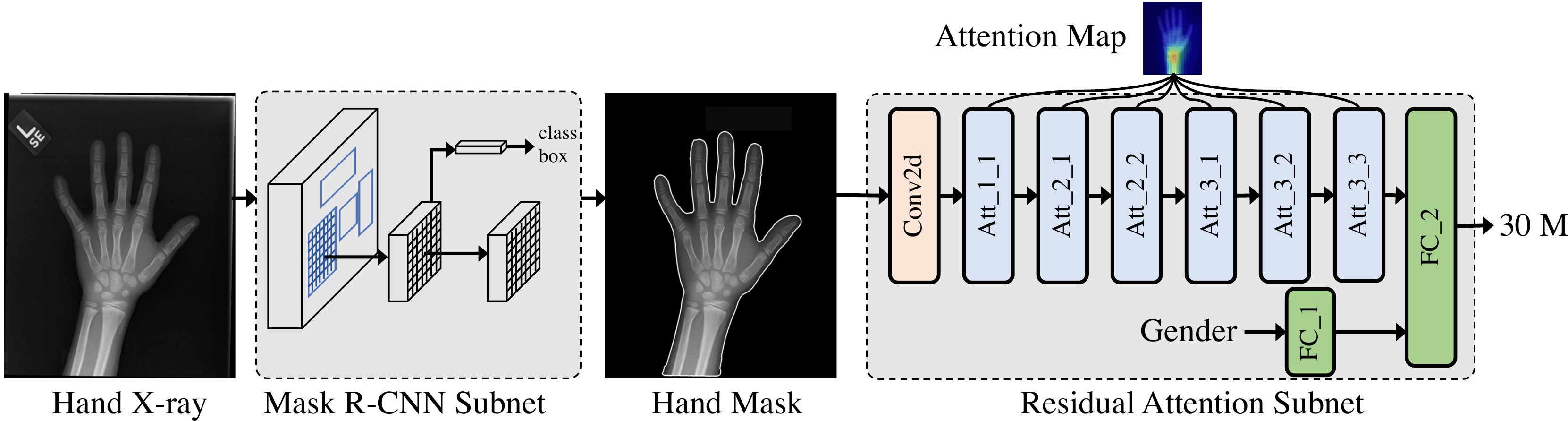} 
\end{center}
~\vspace{-2.0em}
\caption{An overview of our hand X-ray image analysis network for bone age assessment (Note that the residual units are ignored). The network reads hand X-ray images and produces prediction age and mask for the hand.}
\label{fig:overview} 
~\vspace{-1.5em}
\end{figure*}
To address this issue, in this work, we present a unified deep learning network for simultaneous hand segmentation and bone age assessment. Our method consists of two subnets: a Mask R-CNN subnet for pixelwise hand segmentation and a residual attention network for hand bone age assessment. The Mask R-CNN subnet is responsible for segmenting the hands from X-ray images, thereby avoiding the distractions of other objects. The subsequent residual attention network is dedicated to the task of bone age assessment, which leverages attention to force our network to focus on important bone regions. While a similar pipeline is adopted in~\cite{iglovikov2017pediatric}, the segmentation and bone age assessment network are isolated from each other. By contrast, in our method, these two subnets form two steps in a unified framework, which can be trained end-to-end.


In summary, We design a unified deep learning based framework for bone age assessment. It is able to focus on important bone regions. This is achieved by two steps. First, the Mask R-CNN subnet extract hand regions from the X-ray images to remove other objects, so as to avoid their distractions. Second, a residual attention subnet is employed to force the network to automatically attend to important regions. We also evaluate our method on a large bone age assessment dataset, which demonstrates that our design is indeed essential for accurate prediction.


\section{METHODOLOGY}

In this section, we present our approach for bone age estimation. As is illustrated in Fig.~\ref{fig:overview}, our model first use Mask R-CNN \cite{he2017mask} to mask out non-relevant pixels such as those belonging to the extra objects. It then uses a residual attention based network~\cite{wang2017residual} to perform estimation on the masked image, yielding the predicted bone age in months, which is a single number.

\subsection{Network Architecture}
\textbf{Mask R-CNN for hand segmentation:} As is mentioned in Sec.~\ref{sec:intro}, the noisy background may cause undue attention to other parts of X-ray images. Therefore, we employ a segmentation network to remove the distractions. The current state-of-the-art image instance segmentation system is Mask R-CNN \cite{he2017mask}. Mask R-CNN is based on Faster R-CNN \cite{ren2015faster}, but adds a parallel branch for predicting object masks. Mask R-CNN's mask branch corrects for misalignment in ROI proposals using a RoIAlign layer, which significantly increase accuracy.




\textbf{Residual attention network for bone age assessment:} We then use the residual attention network for bone age assessment. Inspired by the practices of the domain experts, the attention mechanism has gained popularity recently to guide all kinds of neural networks to salient features. Attention weighs parts of the input differently to strengthen or diminish features in main network layers. This is commonly accomplished by having a separate branch that calculates attention and is later incorporated back into the main branch with some weighing function. 


As is illustrated in Fig.~\ref{fig:overview}, our residual attention subnet (residual units are ignored for simplicity) is composed of a convolution layer, followed by 6 residual attention modules~\cite{wang2017residual}. Each attention module has a trunk branch $\mathcal{T}$ and soft mask branch $\mathcal{M}$ inside it. Given the input feature map $\textbf{x}$, the attention module generates the trunk branch output $\mathcal{T}(\textbf{x})$ and soft mask map $\mathcal{M}(\textbf{x})$. The trunk branch contains only two few residual blocks and acts as a shortcut for data flow. Afterward, the attention mask $\mathcal{M}(\textbf{x})$ is applied to the trunk branch as a multiplier, generating the attended feature map $\textbf{x}^{\prime}$.  
\begin{align}
\textbf{x}^{\prime}=\mathcal{T}(\textbf{x}) \odot (1 + \mathcal{M}(\textbf{x})),
\label{eq: attention}
\end{align}
where $\odot$ denotes Hadamard product.

While the attention mechanism is also used in~\cite{lee2017fully}, it is only used to visually show significant parts in the images for bone age prediction. Instead, attention in this work is integrated into the network. Thus, our network is guided by attention to focus on important bone regions during training, so that more precise estimations can be made. We also added a gender input to the last fully connected layer that was then concatenated to the input for the final fully connected layer. 

\subsection{Loss function}
Formally, the loss function $\mathcal{L}$ during training is a combination of multiple tasks:
\begin{align}
&\mathcal{L}=\mathcal{L}_{cls}+\mathcal{L}_{box}+\mathcal{L}_{mask}+\mathcal{L}_{reg},\\
&\mathcal{L}_{reg}=\frac{1}{N}\sum_{i=1}^{N}(\hat{y}_i-y_i)^2,
\label{eq: loss_func}
\end{align}
where $\mathcal{L}_{cls}$, $\mathcal{L}_{box}$, and $\mathcal{L}_{mask}$ are the classification loss, bounding-box loss, and per-pixel hand segmentation loss, respectively, which are identical as defined in~\cite{he2017mask}. $\mathcal{L}_{reg}$ is the standard L1 loss, which is the mean absolute error (MAE) between the predicted age $\hat{y}_i$ for the $i^{th} (i=1,2,...,N)$ X-ray image and the ground truth $y_i$.


\section{Experiments}
\textbf{Dataset, preprocessing, and Evaluation Metrics.} Our framework was evaluated on the RSNA pediatric bone age dataset, which includes approximately 12,500 labeled images. All images have a gender associated with them. We used a 90\%/10\% training/validation split. The hand masks in the training set were segmented using the Canny edge detector on histogram normalized images. The MAE (Eq.~\ref{eq: loss_func}) was also used for evaluation. The network was trained using Nesterov SGD with a weight decay of 0.0001, momentum of 0.9, and initial learning rate of 0.01. The learning rate was divided by 10 automatically when validation loss plateaued for 5 epochs. Standard data augmentation transformations (random cropping, resizing, rotation, and mirroring) were used.  

\textbf{Comparison with baseline models.} We compare our model with a baseline model with VGG-16 and the state of the art~\cite{iglovikov2017pediatric}, which report results with two different settings: the first is obtained on whole hands with a combination of U-Net and VGG-style neural network and the second result is obtained by heavy ensembling. In the second setting, the X-ray images were first registered so that the hands are in the same direction. And three networks were trained with different regions of the images. The final predictions were generated by ensembling the results together. The results are shown in Table~\ref{tab:results}. The error of our network was 7.38 months. This is comparable with their MAE of 8.08 months for the whole hand, especially considering that our model was trained without any registration of images. It also slightly exceeds their MAE of 7.52 months for a multi-model ensemble while only using a single network for regression. 

\begin{table}[h]
\caption{Comparison of the proposed method with the state of the art~\cite{iglovikov2017pediatric}.}
\begin{center}
\label{tab:results}
\renewcommand{\arraystretch}{1}
\begin{tabular}{ c c }\toprule
Model & MAE (months) \\ \midrule
        VGG-16 & 14.0  \\ 
        Iglovikov et al.~\cite{iglovikov2017pediatric} on whole hand & 8.08 \\
        Iglovikov et al.~\cite{iglovikov2017pediatric} with ensemble  & 7.52 \\
        Ours & \textbf{7.38}  \\
\bottomrule
\end{tabular}
\end{center}
\end{table}

\begin{figure}[h]
    \centering
    \includegraphics[width=0.95\linewidth]{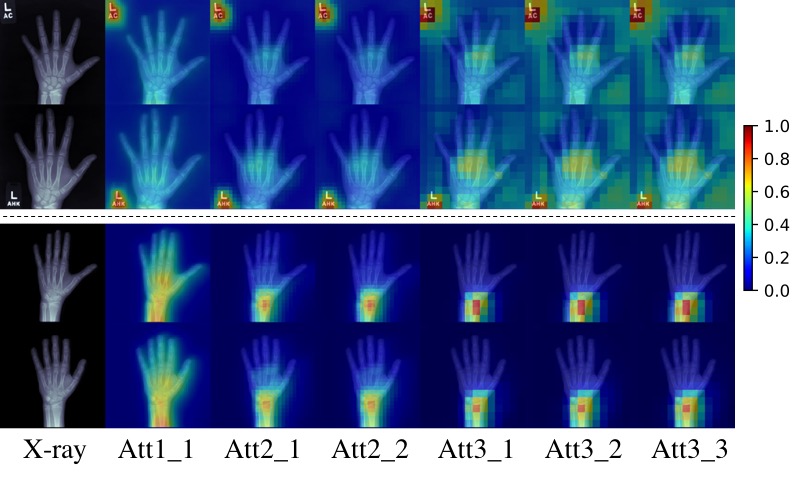}
    \caption{Generated attention maps with and without Mask-RCNN subnet. Top two rows show the input X-ray images and the generated attention maps without Mask-RCNN subnet. Bottom two rows show the segmented hand and the attention maps with the proposed method.}
    \label{fig:AverageAttention}
\end{figure}

\begin{figure}[h]
    \centering
    \includegraphics[width=.95\linewidth]{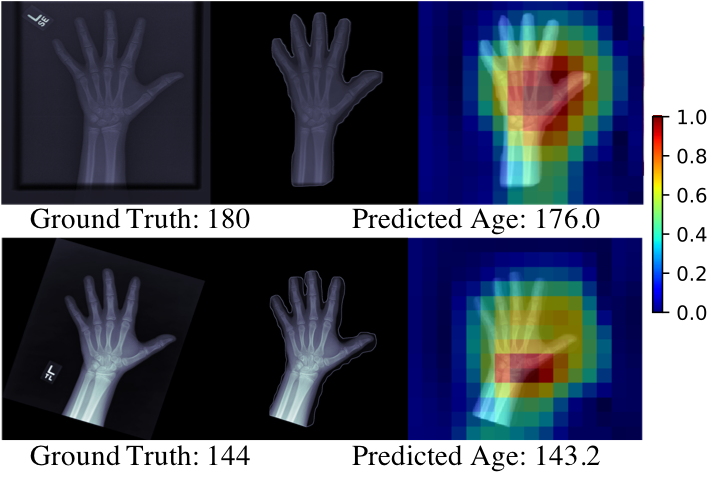}
    \caption{Input X-ray images (left), segmented hands (middle), and highest attention responses (right) from the presented method.}
    \label{fig:NoisyImage}
\vspace{-1.0em}
\end{figure}

 
\textbf{Evaluation of Mask R-CNN subnet}. We then evaluate the Mask R-CNN subnet. Running the final network without Mask R-CNN subnet on images with tags results in an MAE of 12 months, which performs much worse than the presented framework. We also show the attention maps from different attention modules. We take masks from before the sigmoid in each attention module's soft mask branches and map the values with a heatmap. These individual attention maps are shown in Fig.~\ref{fig:AverageAttention}. When there is no Mask R-CNN subnet (top two rows of Fig.~\ref{fig:AverageAttention}), high attention is generated on X-ray tags in all the attention modules. On the contrary, our model does decently well on these images. Figure~\ref{fig:NoisyImage} shows noisy images with background borders and tags (left), the segmented hands (middle), and the attention maps with the highest responses (right) in all 5 attention modules. Mask R-CNN isolates the hand relatively well, and the important regions related to bone age is then correctly focused by the attention mechanism of the residual attention network. 

\textbf{Evaluation of the residual attention subnet.} We can also determine what the network marks as important from the generated attention maps. From Fig.~\ref{fig:AverageAttention} (bottom 2 rows), it can be seen that earlier attention modules focus on the entire hand, while later attention focuses on parts of the hand more pertinent to deciding bone age. The highest average attention was on the carpals in Att3\_1, Att3\_2, and Att3\_3. This is consistent with previous work indicating that the carpals are important in skeletal maturity assessment for infants and toddlers \cite{iglovikov2017pediatric}. In other attention modules, there is more attention on the metacarpals and phalanges than the carpals. We also quantitatively evaluate the attention in our residual attention network. The masks of attention modules Att1\_1, Att2\_1, Att2\_2, and Att3\_1 are sufficiently unique enough that their removal would likely be highly detrimental. Instead, we remove Att3\_2 and Att3\_3 since their masks are visually similar to Att3\_1. The results are shown in Table \ref{tab:exp}. By removing the attention from Att3\_2 and Att3\_3, the MAE increases by 1.69, demonstrating that attention is indeed essential for the bone age assessment.


\begin{table}[h]
\vspace{-0.5em}
\caption{MAE of our method, with or without attention in Att3\_2 and Att3\_3.}
\begin{center}
\label{tab:exp}
\renewcommand{\arraystretch}{1}
\begin{tabular}{ c c }\toprule
Model & MAE (months) \\ \midrule
        Without Att3\_2 and Att3\_3 & 9.07 \\
        Ours & \textbf{7.38}  \\
\bottomrule
\end{tabular}
\end{center}
\vspace{-1.0em}
\end{table}



\textbf{Evaluation of age input.} Finally, we evaluate the importance of age for the prediction. To do this, a separate network exactly the same as in Fig. \ref{fig:overview} except without the gender input was trained. This network attained a mean absolute error of 10 months, which is more than a month higher than the network with a gender input. Therefore, we conclude that gender is an important factor for the network to determine bone age accurately.

\section{CONCLUSIONS}
In this study, we investigate building a deep learning based pipeline for automatic bone age assessment. Inspired by the clinical workflow (Tanner-Whitehouse) of bone age assessment, we build a unified network for age assessment. More specifically, it consists of two subnets: the Mask R-CNN subnet segment hands from the image to remove the distractions from backgrounds, based on which the residual attention subnet generate the final prediction and the visual supports. In this way, our network is more robust to noises.

\textbf{Acknowledgement.} The work received supports from Shenzhen Municipal Government under the grant KQTD 2016112809330877.


\bibliographystyle{IEEEbib}
\bibliography{refs}

\end{document}